# Investigating Bi-LSTM and CRF with POS Tag Embedding for Indonesian Named Entity Tagger

Devin Hoesen
Prosa Solusi Cerdas
Bandung, Indonesia
devin.hoesen@prosa.ai

Ayu Purwarianti
School of Electrical Engineering and Informatics
Institut Teknologi Bandung
ayu@stei.itb.ac.id

*Abstract*—Researches on Indonesian named entity (NE) tagger have been conducted since years ago. However, most did not use deep learning and instead employed traditional machine learning algorithms such as association rule, support vector machine, random forest, naïve bayes, etc. In those researches, word lists as gazetteers or clue words were provided to enhance the accuracy. Here, we attempt to employ deep learning in our Indonesian NE tagger. We use long short-term memory (LSTM) as the topology since it is the state-of-the-art of NE tagger. By using LSTM, we do not need a word list in order to enhance the accuracy. Basically, there are two main things that we investigate. The first is the output layer of the network: Softmax vs conditional random field (CRF). The second is the usage of part of speech (POS) tag embedding input layer. Using 8400 sentences as the training data and 97 sentences as the evaluation data, we find that using POS tag embedding as additional input improves the performance of our Indonesian NE tagger. As for the comparison between Softmax and CRF, we find that both architectures have a weakness in classifying an NE tag.

## I. INTRODUCTION

Named entity (NE) tagger is an important task in natural language processing, especially in information extraction and semantic role labeling. This is also applied for Indonesian language. Thus, there are already several researches for Indonesian NE tagger [1, 2, 3, 4, 5, 6]. Most researches on Indonesian NER employed traditional machine learning algorithms such as association rule [1], ensemble learning [4], and support vector machine (SVM) [2, 3, 5]. The problem of these researches is the features. Most researches depend on word list feature whether it is a gazetteer or a clue word list. Thus it will be difficult for a new NE type to have good classification accuracy since it needs a pre-defined word list.

Research [6] employed deep learning algorithm for Indonesian NE tagger by comparing hybrid Bi-LSTM-CNN with other topologies on top of Bi-LSTM. It followed the state-of-the-art NE tagger using long short-term memory (LSTM) technique researched in [7]. However, the former differed with the latter in that the former didn't use conditional random field (CRF) as the output layer. Research [7], rather than only using softmax in the output layer, employed LSTM with linear-chain CRF for several languages (English, Germany, Dutch, and Spanish) and achieved the highest F1 score for Germany and Spanish compared to other related researches.

Since there is no research on Indonesian NE Tagger using deep learning and CRF as its output layer, we try to investigate the usage of LSTM and CRF for Indonesian NE Tagger. We define several NE types to evaluate the LSTM, not only common NEs such as people (PER) and location (LOC), but also uncommon ones such as event (EVT), products/brands (IND), and food and beverages (FNB). The list of NE labels is shown in Table 1.

TABLE I. TYPE OF NAMED ENTITY USED IN THE RESEARCH

| NE Tag | Explanation |
|---|---|
| PER | Name of people |
| LOC | Name of location |
| IND | Name of products and brands |
| EVT | Name of events |
| FNB | Food and beverage name |

In the LSTM-based Indonesian NE Tagger, we also investigate the usage of POS tag embedding layer as an additional input layer to word and character embedding input layer.

## II. BI-LSTM-CRF MODEL FOR NE TAGGER

### A. LSTM

Recurrent neural network (RNN) are neural network that is claimed to be more suitable for temporal sequence data [8]. In this type of neural network, instead of encoding temporal representation into the input features (e.g. by using sliding window over input features), it is encoded as the effect it has on the processing network by employing some memory units. The network takes a sequence of input vectors and outputs another sequence of vectors that gives information about the inputs at every time step.

The network, in theory, can learn long temporal dependencies of the inputs. However, in practice, the network tend to give more weight to its most recent inputs [9]. LSTM tries to overcome the problem by employing some functions that decide whether some parts of information must be remembered or forgotten [10]. Specifically, given input vectors ($\mathbf{x}_1, \mathbf{x}_2, \ldots, \mathbf{x}_n$), our LSTMs compute their state sequence ($\mathbf{h}_1, \mathbf{h}_2, \ldots, \mathbf{h}_n$) at time-step $t$ by following these equations:

$$\mathbf{i}_t = \sigma(\mathbf{W}_i \mathbf{x}_t + \mathbf{U}_i \mathbf{h}_{t-1} + \mathbf{b}_i)$$

$$\mathbf{f}_t = \sigma(\mathbf{W}_f \mathbf{x}_t + \mathbf{U}_f \mathbf{h}_{t-1} + \mathbf{b}_f)$$

$$\mathbf{c}_t = \mathbf{f}_t \circ \mathbf{c}_{t-1} + \mathbf{i}_t \circ \tanh(\mathbf{W}_c \mathbf{x}_t + \mathbf{U}_c \mathbf{h}_{t-1} + \mathbf{b}_c)$$

$$o_t = \sigma(\mathbf{W}_o \mathbf{x}_t + \mathbf{U}_o \mathbf{h}_{t-1} + \mathbf{b}_o)$$
$$\mathbf{h}_t = \mathbf{o}_t \circ \tanh(\mathbf{c}_t)$$

where $\sigma$ is component-wise logistic function, $\circ$ is component-wise (Hadamard) product, **W** are weights for **x**, **U** are weights for **h**, and **b** is bias value. Function **i**, **f**, and **c** consecutively denote input gate, forget gate, and cell's state function. Subscripts for matrix **W**, **U**, and **b** denote which gate the matrix belongs to.

### B. Bi-LSTM-CRF NE Tagger

Unidirectional forward LSTM layer only remembers and/or forget past dependencies. In NE tagging, both past and future dependencies can give information about the current NE. Those dependencies can be captured by Bidirectional LSTM (Bi-LSTM) layer first proposed in [11]. In Bi-LSTM "layer", there are two layers of LSTM cells; one layer is forward LSTM layer to capture past dependencies, and another layer is backward LSTM layer to capture future dependencies. An example of Bi-LSTM architecture is shown in Figure 1.

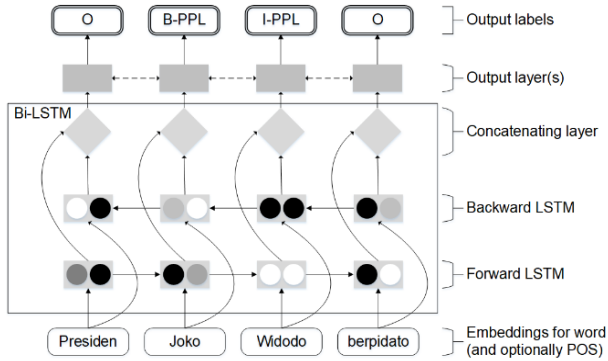

Figure 1. Architecture of the NE tagger with Bi-LSTM. The dashed line in the output layer illustrates an optional chain CRF for outputting the labels

In order to capture strong interdependence between NE labels, conditional random field (CRF) in form of linear chain CRF can be employed together with Bi-LSTM. First proposed in [12], CRF is suitable of NE tagging because the tagging has several hard constraints (e.g. I-FNB cannot follows both O and B-EVT). Instead of assuming that every resulting NE label is independent of each other, CRF assumes that it is globally (i.e. for the whole NE tag sequence) interdependent. Together with Bi-LSTM, linear chain CRF can encourage the model to produce the valid **sequence** of NE tags rather than only valid independent class of NE tag [7].

### C. Indonesian NE Tagger Network Architecture

The network architecture used in this research is similar to the one in [7]. It is illustrated in Figure 1. Different with [7], our architecture takes on word features and POS embedding as its input. The word features consist of word embedding and its character-to-word (C2W) embedding [13]. For the word embedding, we use pre-trained word2vec's skip-gram embedding discussed in the previous sub-section.

The illustration for C2W embedding is shown in Figure 2. We use 25 units for each of the forward and backward LSTM layer as described in [7] resulting in 50-dimension embedding vector for each word. However, we limit the permitted word's characters to be only alphanumeric lower-case characters. All word's characters must be lower-cased because we want the letter case to not affect the tagging result. Furthermore, all numeric characters are normalized to just '0' (zero) character. Symbols and other non-alphanumeric characters are mapped to "<UNK>" special character.

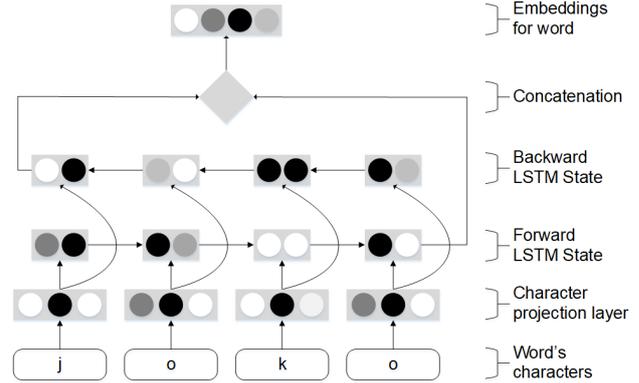

Figure 2. Illustration of character to word (C2W) embeddings for word "joko"

In addition to word features, we also want to evaluate the effect of POS embeddings to NE recognition, thus we add an optional projection (embedding) layer for a word's POS. The projection layer produces 25-dimension POS embedding vectors to be fed to the main network.

The inputs are fed to the main network that consists of a bidirectional LSTM layer and a fully connected layer on top of it. Each of the forward and backward layer in the bidirectional layer has 100 LSTM units. The fully connected layer also has 100 hidden units with "tanh" activation.

For the output layer(s), we want to evaluate the effectiveness of the linear chain CRF to recognize NE. Thus, we have architectures that have linear chain CRF applied on top of a fully-connected linear output layer and architectures that only have fully-connected softmax output layer. Both types of output layer(s) rest on top of the fully-connected "tanh" layer described before.

## III. EXPERIMENTS

### A. Training and Evaluation Data

The training data comprises 8,400 sentences while the evaluation data comprises 97 sentences. Both are articles extracted from some Indonesian news websites. The sentences are manually tokenized so that each word, each symbol, and each number become separated token. For currency, if the currency symbol is written with no space from its value, they will be regarded as one token. If the currency symbol is written separately from its value, each will be regarded as a token. Moreover, if the separated currency symbol contains symbol of the country that the currency belongs to, each of the country and the currency symbol will become separated token.

Each token is then annotated with its POS tag and NE tag. There are 26 POS tag classes as described by Indonesian Association of Computational Linguistics (INACL) [14]. The POS tags are explained in Table 2. On the other hand, there are 5 NE labels used in this research. The labels and their own explanation have been shown in Table 1. Because named entities can consist of several tokens, we use the IOB (Inside, Outside, Beginning) tagging format, where every token is labeled B-*label* if it is a beginning of an NE, I-*label* if it is an NE token but not the beginning, and O if it belongs to no NE.

In the training phase, we use pre-trained word embedding vectors. The texts for training the embedding vectors were taken from some Indonesian news websites, i.e. *Kompas, MetroTV News, Republika*, and *Tempo*, ranged from 2008 until 2016. After automatic removing of punctuation and number conversion to text, the text corpus contained 24,469,110 lower-case sentences and 451,171,582 words. Word2vec's 100-dimension skip-gram vectors was trained using the texts with context window of ±5 and negative sampling of 5 samples. Limiting words that occurs at least 11 times, the vocabulary for the vectors contained 253,849 unique words.

*B. Experiment Results*

Overall, for the experiment, there are four architectures that are built by combinating the existence of POS embedding input layer and whether the output layer is a linear chain CRF or softmax layer. The four architectures are named as follows.

1. **CRF**, architecture that uses linear chain CRF output layer without POS embedding input.
2. **CRF-POS**, architecture that uses linear chain CRF output layer with POS embedding input.
3. **Softmax**, architecture that uses softmax output layer without POS embedding input.
4. **Softmax-POS**, architecture that uses softmax output layer with POS embedding input.

We use F1 score as the metric for each class to compare the experimental results. An NE is correct if reference. The overall F1 score for each architecture is shown in Figure 3. Even though **Softmax-POS** shows the highest F1 score, but if we see it in detail, the results of **CRF-POS** aren't counted fairly because in the F1 calculation, we use precise matching between the reference and the prediction results. It turns out that the precise matching might not be a good metrics for our Indonesian NE tagger since there are cases where the **CRF-POS** is able to extract a portion of correct terms. For example, in sentence "*Dodee Paidang memberikan promo happy hour cuma dengan rp 35 ribuan aja loh!*", the words "*promo happy hour*" are tagged manually as EVT. Here, the **Softmax** is unable to extract any words of it and classify all three words as OTHER, while the **CRF-POS** is able to extract "*happy hour*" as EVT. But, since the correct reference is "*promo happy hour*", then the score for **CRF-POS** is 0 for this case.

TABLE II. POS TAGS AS LISTED BY INACL

| POS Tag | Explanation |
|---------|-------------|
| NNO | Noun |
| NNP | Proper noun |
| PRN | Pronoun |
| PRR | Relative pronoun |
| PRI | Interrogative pronoun |
| PRK | Cliticized pronoun |
| ADJ | Adjective |
| VBI | Intransitive verb |
| VBT | Transitive verb |
| VBP | Passive verb |
| VBL | Linking verb |
| VBE | Existential verb |
| ADV | Modal adverb |
| ADK | Time adverb |
| NEG | Negation |
| CCN | Coordinative conjunction |
| CSN | Subordinative conjunction |
| PPO | Preposition |
| INT | Interjection |
| KUA | Quantifier |
| NUM | Numeral |
| ART | Article |
| PAR | Particle |
| UNS | Unit symbol |
| $$$ | Currency |
| SYM | Character symbol |

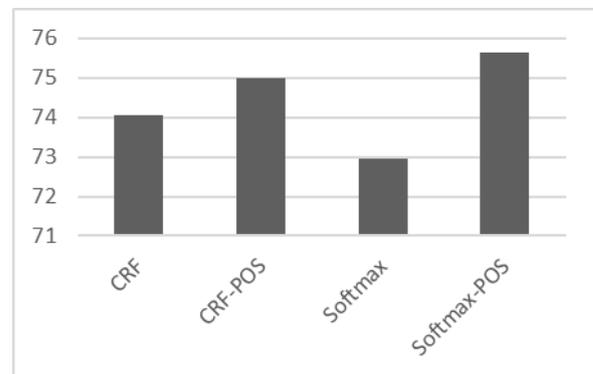

Figure 3. The Overall F1 Score

Figure 3 also shows that POS tag embedding layer is able to enhance the F1 score for both architecture (using softmax and CRF). The POS tag information gives additional clue on the appropriate NE tag. For example, in sentence "*akun @henjiwong mencatatkan lebih dari 2.400 post **Instagram** dengan 46.100 followers.*", the word "*Instagram*" has the correct NE type of IND. Here, the architecture without POS tag information get the phrase "*post Instagram*" as IND, while the one with POS tag information correctly get only the "*Instagram*" as IND.

We also try to see the performance of each NE class. The complete F1 score for each class is shown in Figure 4. The F1 score for each class is rather similar for both CRF and softmax architectures. The detailed results are different for each class. For EVT and FNB, softmax architectures have higher F1 score compared to CRF ones.

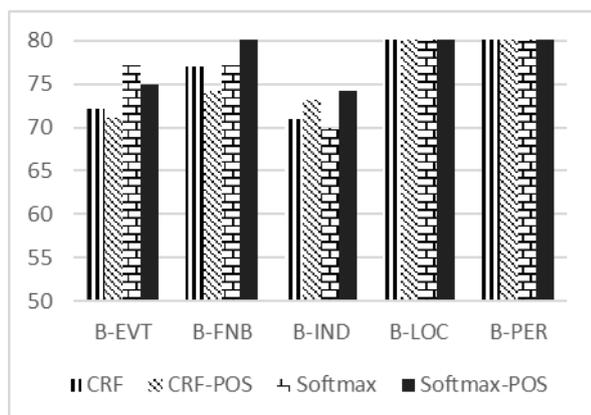

Figure 4. F1 Score for all NE Tag

There is a tendency that CRF will take surrounding words with noun POS-tag as EVT or FNB classes' part, since many common nouns became part of the classes. For example, in sentence "*Ada 5 pilihan menu snack dan semua paket yang sudah termasuk thai tea/coffee di dalamnya!*", words that actually are in FNB class are "*snack*", "*thai tea*", and "*coffee*". The **Softmax-POS** and the **CRF** architecture give the correct results. The POS information gives another clue for the softmax and can change the NE tag for "*snack*" from OTHER class (for "Softmax" architecture) to B-FNB (for "Softmax-POS" architecture). This situation is inversed in CRF architectures. The one without POS tag can classify "*snack*" as B-FNB. On the other hand, the one with POS tag misclassify the concerned FNB as "*menu snack*" where "*menu*" becomes the B-FNB and the "*snack*" becomes I-FNB.

For the LOC, PER and IND class, Figure 4 shows that CRF architectures have similar F1 score to the softmax ones. But in the detailed results, the CRFs show better recognition than the softmaxes. For example, in sentence "*yap, adalah warung bakso kumis permai vi yang jadi korban berita hoax kali ini*", the words "*warung bakso kumis permai vi*" are manually tagged as IND. The CRFs successfully aggregate them into one NE tag although it is an incorrect tag of LOC, while the softmaxes misclassify them into LOC, FNB and IND.

## IV. CONCLUSION

We have conducted experiments on Indonesian NE Taggers with Bi-LSTM architecture. The experiment results on 5 NE tags give two conclusions. First, the POS tag embedding additional input gives higher F1 score for both architectures with CRF and softmax as their respective output layer. Second, the best architecture between CRF and softmax depends on the NE tag.

## V. ACKNOWLEDGEMENT

This work was funded by the Indonesian research program 'PENELITIAN TERAPAN UNGGULAN PERGURUAN TINGGI' with title 'Sistem Cerdas Pemantau Perilaku Penggunaan Gadget di Kalangan Remaja Menggunakan Teknik Pembelajaran Mesin' (intelligent system for monitoring gadget usage behavior among teenagers using machine learning). We also would like to thank you other parties who contributed in this experiment or during the process of working for this article.